\documentclass{article}
\usepackage{arxiv}
\usepackage{scrextend}
\usepackage{cite}
\usepackage{amsmath,amssymb,amsfonts}
\usepackage{algorithmic}
\usepackage{textcomp}
\usepackage[utf8]{inputenc}
\usepackage[T1]{fontenc}
\usepackage{hyperref}
\usepackage{url}
\usepackage{booktabs}
\usepackage{amsfonts}
\usepackage{nicefrac}
\usepackage{microtype}
\usepackage{graphicx}
\usepackage{xcolor}
\usepackage{svg}
\usepackage{amsmath}
\usepackage{float}
\usepackage{subfig}
\usepackage{enumitem}
\usepackage{hyperref}

\title{Teamwork under extreme uncertainty:\\AI for Pok\'emon ranks 33rd in the world}

\author{ {\hspace{1mm}Nicholas R. Sarantinos} \\
	University of Cambridge \\
	\texttt{nrs38@cantab.ac.uk}
}

\date{}

\renewcommand{\undertitle}{}

\begin{document}

\maketitle

\begin{abstract}
The highest grossing media franchise of all times, with over \$90 billion in total revenue, is Pok\'emon. The video games belong to the class of Japanese Role Playing Games (J-RPG). Developing a powerful AI agent for these games is very hard because they present big challenges to MinMax, Monte Carlo Tree Search and statistical Machine Learning, as they are vastly different from the well explored in AI literature games. An AI agent for one of these games means significant progress in AI agents for the entire class. Further, the key principles of such work can hopefully inspire approaches to several domains that require excellent teamwork under conditions of extreme uncertainty, including managing a team of doctors, robots or employees in an ever changing environment, like a pandemic stricken region or a war-zone. In this paper we first explain the mechanics of the game and we perform a game analysis. We continue by proposing unique AI algorithms based on our understanding that the two biggest challenges in the game are keeping a balanced team and dealing with three sources of uncertainty. Later on, we describe why evaluating the performance of such agents is challenging and we present the results of our approach. Our AI agent performed significantly better than all previous attempts and peaked at the 33rd place in the world, in one of the most popular battle formats, while running on only 4 single socket servers.
\end{abstract}

\section{Introduction}
\label{sec:introduction}

Since the early days of Artificial Intelligence, a lot of research was focused on developing intelligent agents for games. Despite the significant amount of work done in deterministic and perfect information games, including Chess \cite{b53} and GO \cite{b1}, there is limited research in games like Pok\'emon, that combine a significant random factor with imperfect information, outside of Poker.

\begin{figure}
    \centering
    \subfloat[A player is exploring the vast game world.]{{\includegraphics[scale=0.255]{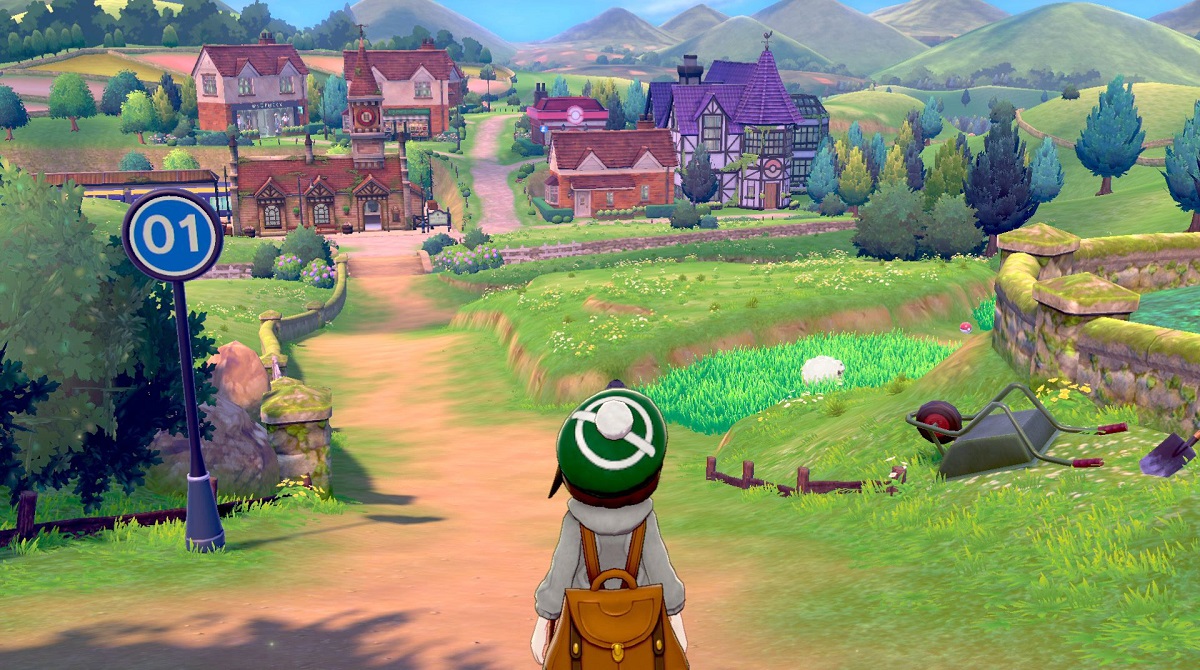}}}
\vspace{1mm}
    \subfloat[Two Pok\'emon battle against each other. This work proposes an AI agent for battles like this one.]{{\includegraphics[scale=0.255]{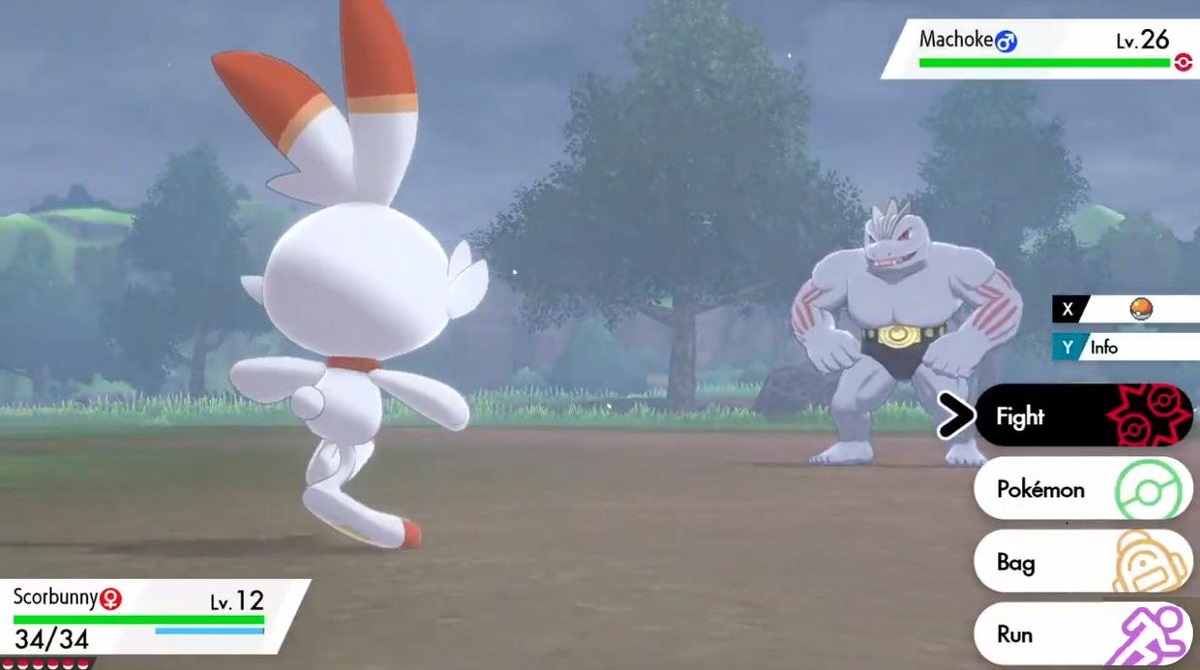}}}
    \caption{Pok\'emon Sword and Pok\'emon Shield, the first 8th generation core series Pok\'emon games.}
    \label{fig:pokemon_game_images}
\end{figure}

Pok\'emon, the highest grossing media franchise of all times, long ago inspired fans to develop intelligent agents capable of competing with human players \cite{b3} \cite{b4} \cite{b5} \cite{b6}. The franchise began in 1995 with the introduction of the first Pok\'emon games for the Nintendo Game Boy platform. In these games players explore a vast game world, collect, train, evolve and trade creatures known as Pok\'emon as well as battle against other players using a team of Pok\'emon. Since 1995 over 70 Pok\'emon games have been released and together with merchandise, movies, TV shows, cards and more they grossed over \$90 billion in total revenue. The Pok\'emon games can be classified as either core series \cite{b43} or spin-offs \cite{b44}. The differences between the two categories are outside the context of this work and here we focus only on the former. On 18 November 2022 the first 9th generation core series games, Pok\'emon Scarlet and Pok\'emon Violet, were released. When this work took place there were 8 generations of games and the latest core series games were Pok\'emon Sword and Pok\'emon Shield. Screenshots of these games are in Figure~\ref{fig:pokemon_game_images}. While every new generation brings changes, the games always allowed two players to connect their consoles locally and battle against each other. Since 2006, players can do the same via Internet. In 2009 and every year afterwards, the game publisher, Nintendo, organizes the ``Video Game Championships (VGC)'' where the top players across the globe are invited to battle against each other for the title of World Champion. This and other events led to the emergence of professional Pok\'emon players. Nintendo designed the games to be easy for young kids to play but difficult for professional players to master as the battle mechanics are very complicated. Numerous fan made websites provide extensive analysis of the game mechanics, in order to aid professional and amateur players \cite{b7} \cite{b8} \cite{b9}.

Developing a powerful AI agent for pok\'emon battles can inspire all sorts of applications that require team management under conditions of extreme uncertainty, including managing a team of doctors, robots or employees in an ever changing environment, like a pandemic stricken region or a war-zone. Further, such an agent can be used with tweaks in numerous other popular Japanese Role Playing Games (J-RPG), like Final Fantasy, as many of these games share similar battle mechanics and victory depends on teamwork under extreme uncertainty.

There have been several attempts by others to develop AI agents for Pok\'emon \cite{b39} \cite{b35} \cite{b40} \cite{b3} \cite{b4} \cite{b37} \cite{b36} \cite{b38}. These pioneer attempts gave us a better understanding of the challenges involved. They mostly relied on work done in Chess or GO or tackled the entire problem with a reinforcement learning algorithm and did not invest the enormous amount of time and effort required to invent the new approaches Pok\'emon requires, as it is a vastly different problem.

We will now describe how this paper is organized. In section 2 (``Background''), we explain in adequate depth the mechanics of the game from the perspective of an AI researcher, we introduce the online battling platform our work relies on, we present the work by others so far and finally we observe that our work can inspire approaches in many real world problems. In section 3 (``Game analysis from an AI perspective''), we perform a game analysis that vastly expands our understanding of the problem. In section 4 (``Methods of this paper'') we propose unique AI algorithms based on our understanding that the two biggest challenges in the game are keeping a balanced team and dealing with three different sources of uncertainty. In section 5 (``Experiments''), we describe why evaluating the performance of Pok\'emon AI agents is challenging and we present the results of our unique AI algorithms. The results place our AI agent above all attempts by others so far by quite a margin. It peaked at the 33rd place in the world, in one of the most popular battle formats. To the best of our knowledge, there is no other AI agent for Pok\'emon that has managed to reach the top 500 players in the world in any battle format.

\section{Background}
\label{sec:background}

\subsection{The mechanics of Pok\'emon battles}

In this section we describe the mechanics behind Pok\'emon battles. We first describe how battles are played, then we proceed to describe the properties of the creatures known as Pok\'emon and finally we describe the format of the battle states. We will not describe how the game world is explored as well as how Pok\'emon are collected, trained, evolved and traded because these mechanics are outside the context of this work. The battles we focus on are known as competitive because the only goal is to win, unlike other battles in the game where the goal is to train a Pok\'emon or capture a wild one. The following apply to the core series games of generations 6 to 8. The reader can find more in-depth information at online sites \cite{b7} \cite{b8} \cite{b9}.

There are two battling sides and one, two or three active Pok\'emon at every side at every moment. Every side has also reserve Pok\'emon that can be called to replace active ones at any point. Pok\'emon battles are categorized as Single, Double or Triple, depending on the number of active Pok\'emon per side. The battle sides usually have the same number of active Pok\'emon. This work focuses on Single battles but can be expanded to support the other two categories too. The total number of Pok\'emon per team, both active and reserve, is defined by the rules of the specific battle and can never exceed six. Our work focuses on battles with the maximum number of Pok\'emon per team.

The games use a turn based battling system. At the start of every turn the two sides must select their actions simultaneously, without knowing the selection of the opponent. Actions are only executed after both sides make their selections. We describe the order of execution below. These actions, as seen in Figure~\ref{fig:pokemon_game_images} (b), can be: 

\begin{enumerate}
\item Make an active Pok\'emon use a move. Every Pok\'emon can use at most one move per turn.
\item Switch/replace an active Pok\'emon with a reserve. Every Pok\'emon can be switched with another at most once per turn.
\item Use an item from the bag \cite{b48}. There are several kinds of items stored there and their use is to aid the exploration of the game world as well as the collection and training of Pok\'emon.
\item Flee the battle. This is allowed only if the opponent is a wild Pok\'emon and not another player.
\end{enumerate}

In competitive battles, which we are focusing, only the first two are possible. Players cannot use their bag in these battles but each Pok\'emon is allowed to hold an item. The order selections are executed depends on game mechanics. Barring rare exceptions, Pok\'emon switching is executed before any move. The order at which moves are executed depends on many of the Pok\'emon properties we describe later on. The three most important are the speed statistic, ability and move effect. Every player knows the properties of their Pok\'emon but not these of the enemy ones. As a result, players have to guess this hidden information in order to determine which Pok\'emon will move first.

If a player switches an active Pok\'emon with another, while their active Pok\'emon is still able to battle, this will cause the newly switched in Pok\'emon to receive a hit upon entering the battle arena, without having the time to attack back, if the opponent chooses to use a move. For this reason switching Pok\'emon without being forced to do so is risky and should ideally be done only if the expected opponent move will cause little damage or if the opponent is also expected to switch Pok\'emon too. 

The winner in the battle is the player that manages to reduce to zero the life of all opponent Pok\'emon first. One has to consider team balance, as a team contains multiple Pok\'emon and one will have to decide if it is best to have one fully healthy or many damaged ones. The goal is not just to damage the enemy as much as possible but to make sure all their Pok\'emon will faint at the end. This is complicated by the vast amount of conditions that affect the battle, as we describe later on. Estimating the value of battle states is incredibly complicated as, on top of taking into account the team balance, the battle conditions and the uncertainty hidden information result in, there is an extraordinary amount of corner cases. Sometimes it is a mistake to reduce the life points of an opponent Pok\'emon, unless they reach zero, as this may activate an ability that doubles its attacking power. In many cases even professional players have heated arguments as to which battle state is preferred.

Having described how battles are played, we will now explain in adequate detail the most important, in the context of this work, properties of Pok\'emon.

\begin{enumerate}

\item \textbf{The species}: Every Pok\'emon belongs to one of the over 800 possible species \cite{b13} \cite{b14}. The species is very important because it impacts indirectly the battle in many ways, as described below.

\item \textbf{The level}: Every Pok\'emon has a level in the range from 1 to 100 \cite{b15}. The level does not impact the battle in any direct way but is significant because it has major indirect impact since, when a Pok\'emon level ups:

\begin{enumerate}
\item Its statistics are increased
\item New moves can be learned
\item It can evolve to another compatible species
\end{enumerate}

\item \textbf{The type(s)}: Every Pok\'emon can have one or two types \cite{b16} out of: Normal, Fire, Water, Grass, Electric, Ice, Fighting, Poison, Ground, Flying, Psychic, Bug, Rock, Ghost, Dragon, Dark, Steel and Fairy. For example Pikachu is Electric type while Charizard is Fire and Flying type. Types are important mainly because they increase or decrease the effectiveness of moves. For example using a Fire type move against a Water type Pok\'emon will halve the damage dealt. Consult a type chart \cite{b17} for details on effectiveness.

\item \textbf{The statistics}: Every Pok\'emon has: Hit Points (HP), Attack, Defense, Special Attack, Special Defense and Speed \cite{b18}. These are determined by a formula \cite{b19} that takes into account the Species, Level, Nature \cite{b20}, Individual Values (IV) \cite{b21} and Effort Values (EV) \cite{b22}. The last three parameters are not explained here because they exist only to impact the statistics and have no other value in battles. Once the formula calculates the value of a statistic, this can be temporally modified by items, abilities, move effects and battle conditions.

\item \textbf{The ability}: Every Pok\'emon can have one ability \cite{b23} out of one to three possible that depend on the species. For example the possible abilities for Pikachu are Static and Lightning Rod while for Charizard are Blaze and Solar Power. The ability impacts directly the battle in many ways as it can modify the statistics, paralyze the opponent, change the effectiveness of the moves and have many other effects. Over 250 abilities exist in total \cite{b45}.

\item \textbf{The item}: Every Pok\'emon can optionally hold one of the over 1,500 items available in the game \cite{b24} \cite{b25}. Examples of items are Leftovers and Choice Band. Items can have numerous effects in battles and are usually used to heal or boost the Pok\'emon that holds them.

\item \textbf{The move(s)}: Every Pok\'emon knows one to four moves \cite{b26}. The pool of possible moves depends on the species and can range from only 1 to over 100. For example Pikachu can learn Volt Tackle and Thunderbolt but not Fire Blast. Every move has six properties: type, category, damage, accuracy, power points (PP) and optionally an effect. A non deterministic formula \cite{b28} takes into account the properties of the move as well as the properties of the attacking and the defending Pok\'emon and determines the damage. The move category is either Physical, Special or Status and determines if the Attack and the Defense or if the Special Attack and the Special Defense statistics will be used in the damage calculation. Power points is the maximum number of times the move can be used in a battle. A few move effects can heal the user. Over 700 moves exist in the games \cite{b27}.

\end{enumerate}

Having described the key properties of the creatures known as Pok\'emon, we will now describe the format of the battle states. As one would expect, due to the hidden information in the game, only the computer running the battle software can fully observe the battle state. Every state consists of:

\begin{enumerate}

\item \textbf{The player teams}: The team of every player is stored in the battle state. As mentioned above, depending on the battle rules, the properties of the opponent's Pok\'emon, including in many cases the species, are unknown unless revealed or inferred using the process of elimination. Every player attempts, to the best of their ability, to infer as much of the hidden information about their opponent's team as possible. These inferred information can be considered to be part of the battle state the player observes.

\item \textbf{The battle conditions} \cite{b29}: These can be classified into three categories, depending on their scope, as we do below. There are other ways to classify them too.

\begin{enumerate}

\item \textbf{The battlefield conditions}: These impact the entire battlefield and usually last a few turns. An example is the change of the weather into rain. This boosts the power of water type moves and makes certain electric type moves to never miss the target. There are about 10 such conditions.

\item \textbf{The battle-side conditions}: These impact only one side of the battlefield. Some last a few turns and others for the entire battle, unless removed by an effect. One example is placing toxic spikes at the enemy side, so all enemies will immediately be poisoned when they are switched in. Another is summoning a tailwind at the ally side so all allies will have their speed statistic doubled for a few turns. There are about 20 such conditions.

\item \textbf{The Pok\'emon conditions}: These impact only individual Pok\'emon. Some last a few turns, others last until the Pok\'emon is switched out and others for the entire battle, unless removed by an effect. One example is changing the status of a Pok\'emon into frozen, burned or paralyzed. Another is boosting or decreasing any of the statistics of a Pok\'emon. It is even possible to change the form of a Pok\'emon mid-battle temporally. There are about 100 such conditions.

\end{enumerate}

\end{enumerate}

\subsection{A simulator and a platform for Pok\'emon battles}

The core series Pok\'emon games have been officially available only in Nintendo platforms, including Nintendo Switch. Until recently, the games made it difficult to play ranked battles against others online. In these battles players are ranked in a global leader-board, based on a ranking system, like professional Chess players. Further, many online features are only available for a year or two after the release of a game, requiring players to keep buying the latest version, almost every year, to keep playing online battles. The passion of Pok\'emon fans along with these issues led to the creation of a fan made Pok\'emon battle simulator, Pok\'emon Showdown.

Pok\'emon Showdown was started in 2011 by Guangcong Luo and is currently open source with over 20,000 commits and over 300 contributors on GitHub \cite{b30}. The project successfully simulates the extremely complicated Pok\'emon battle system and is virtually perfect, barring rare bugs. It does not simulate the exploration of the game world as well as the collection, training, evolution and trading of Pok\'emon. An online platform was created around the simulator that allows players to play ranked battles. It is accessible via a web page ( \href{https://play.pokemonshowdown.com/}{https://play.pokemonshowdown.com/} ). Players are ranked based on their wins and loses using the ELO rating system. The platform also displays the Glicko rating as well as a custom variation of this. Players are matched with an opponent based on their ELO rating. If either of the players asks for it, the platform sets a time limit of a few seconds per decision. This is very strict and an issue for AI agents.

The online platform offers battles in several different formats \cite{b46}. Each battle format has different rules and presents different challenges to players. The most important differences between the most popular battle formats are the following:

\begin{enumerate}

\item \textbf{Team selection}: At the start of every battle, players either select a team of Pok\'emon to use or they are provided with a team of randomly generated Pok\'emon. Building a great Pok\'emon team is no easy task. On the other hand, it is significantly harder for players to master every possible random team than to master a few carefully selected.

\item \textbf{Opponent team preview}: At the start of every battle, the species property of every opponent Pok\'emon is either revealed or kept hidden until the opponent Pok\'emon in question is used at least once in the battle. In general, if the opponent Pok\'emon species property is kept hidden, the battle is considered significantly harder, as players have to make assumptions about this hidden information.

\item \textbf{Number of active Pok\'emon}: As we discussed above, Pok\'emon battles are categorized as Single, Double or Triple, depending on the number of active Pok\'emon per side.

\item \textbf{Number of Pok\'emon per team}: As we discussed above, Pok\'emon teams can have a maximum of six Pok\'emon, including both active and reserve.

\item \textbf{Restrictions}: The rules of the battle format can include several restrictions, including banned Pok\'emon, moves, items, abilities and more.

\item \textbf{Generation}: Every generation of Pok\'emon games differs in the available Pok\'emon, moves, items, abilities as well as battle mechanics. All generations from 1 to 9 are available on the platform.

\end{enumerate}

The two most popular battle formats in the online platform are named ``Random Battles'' and ``OverUsed (OU) Battles'' \cite{b47}. In the former, players are provided with a random team at the start of the battle and there is no opponent team preview. In the later, players select their team and there is opponent team preview. Both of them are Single battles as players have one active Pok\'emon at a time. Further, in both of them players have teams of six Pok\'emon. There are differences between the restrictions of the two battle formats but they will not be discussed here due to space constraints. In this work we focus on the 7th generation of core series Pok\'emon games. We also focus on the format ``Random Battles'' which is the most popular and is considered by many more challenging than the ``OverUsed (OU) Battles'' format. At this paper we propose an algorithm for team selection, making it easy for others to extend our work and develop an AI agent that supports the other very popular format too.

The simulator and the online platform are the two pieces of infrastructure required to make a Pok\'emon AI agent. Without the simulator it is not possible for an AI to employ look-ahead. Nintendo, so far, have not provided an API to interact with their Pok\'emon games programmatically. As a result, the only way an AI agent can play against human players in a Nintendo platform is by analyzing the video captured from the screen of the console. The online platform based on Pok\'emon Showdown solves this problem as it provides a websocket API that can be used by AI agents to receive the stream of battle events and respond back with battle commands.

\subsection{Work by others}

As the Pok\'emon franchise remains as popular as ever, in part thanks to the very successful spin-off game Pok\'emon GO \cite{b49}, there have been a lot of pioneer attempts to create Pok\'emon AI agents. In addition to what we will present below, there are many other attempts that are only available on GitHub and BitBucket as open source projects \cite{b50} \cite{b5} \cite{b6}.

Ho and Ramesh \cite{b35} while students at Stanford University developed one of the first AI agents that relied on MinMax and a handcrafted evaluation function.

Khosla et al. \cite{b40}, also while students at Stanford University, developed an agent that relied on Expectiminimax and compared its performance to MinMax. Instead of using a handcrafted evaluation function, they used one based on TD learning, a form of reinforcement learning. This work is important because it made clear that evaluating battle states in Pok\'emon is very hard and a handcrafted evaluation function is a poor choice.

Kalose et al. \cite{b3}, again while students at Stanford University, developed an AI agent that used a model free approach and relied on Q-Learning. The conclusion was that this approach requires much more effort than Minmax and so far achieved lower performance.

Chen and Lin \cite{b4}, once again while students at Stanford University, used Deep Reinforcement learning and feature space embeddings. The conclusion was that approach did not performed as well as others.

Lee and Togelius \cite{b39} proposed a competition for Pok\'emon AI agents. Their work focused on comparing AI agents by having them battle each other.

Sanchez \cite{b37} did an extensive study on the area. This work attempted to use the approaches Google Deepmind used to master the game of GO \cite{b1} but concluded that it is not appropriate for the game of Pok\'emon. He also proposed further work on reinforcement learning approaches.

Norstr\"{o}m \cite{b36} did another extensive study in the area. This work focused on a simplified form of the 1st generation core series games. It relied on Monte Carlo Tree search but was not evaluated in the online platform based on Pok\'emon Showdown.

Huang and Lee \cite{b38} developed two reinforcement learning agents that were trained in 3.8 million self play battles over the course of six days. Their first AI agent achieved decent performance against human players as it performed on pair with the best work of others. Further, the agent was put to play against the AI agent \cite{b5} but performed worse. Their second AI agent was trained to play with a few selected teams. This AI agent managed to win most battles against the AI agent \cite{b5} but performed worse than the first against human players.

Dotson \cite{b59} presented an AI agent in a website and a YouTube video. The author did not describe his approach in detail. The results reported indicate that the agent is stronger than the others described above. As we will see below, it is not straightforward to compare the performance of AI agents.

To the best of our knowledge, until now no AI agent for Pok\'emon has managed to reach the top 500 players in any battle format.

\subsection{Applications in other domains}

As we describe in section 3 (``Game analysis from an AI perspective''), victory in the game of Pok\'emon depends on managing a team and ensuring excellent teamwork under conditions of extreme uncertainty. This lies at the heart of many significant real world problems that span several domains. Examples include managing a team of doctors in a case of a polytrauma patient with an ever changing condition, managing a team of employees to complete a project in an very dynamic environment like a pandemic stricken region, managing a team of autonomous drones in a rescue operation against a team of adversaries and many more. We hope that the key principles of our work will inspire approaches that solve problems in these domains. Further, this work can be applied to several other extremely popular Japanese Role Playing Games (J-RPG), like Final Fantasy.

\section{Game analysis from an AI perspective}

In this section we describe our key conclusions about the game. These are very important as they serve the basis of the unique approach to AI agent development we present in the next section.

\subsection{How battles can be won}

From our description of the Pok\'emon battle mechanics, it should be clear that great players have memorized a lot of information: they know the properties of the over 800 Pok\'emon species, over 700 moves, over 1,500 items, over 250 abilities as well as the over 100 possible battle conditions of generation 7. Most importantly, great players understand that victory depends on:

\begin{enumerate}

\item \textbf{Keeping the team as balanced as possible} during the battle. Sometimes it is worth to take a short term loss in order to keep the team balanced and capable to beat the opponent's team. For example, it is worth to lose a fully healthy Pok\'emon in order to make sure the only water type in our team remains undamaged and is able to beat the two Fire type in the opponent's team, otherwise, assuming we have no other way to beat these, we will lose the battle. Every player wants to maximize the probability that their team will win the opponent's team. In other words, at every point, they want to be sure that they have at least a way to beat all opponent Pok\'emon with at least one of their Pok\'emon still standing. An AI agent cannot achieve this just by looking ahead with MinMax or Monte Carlo Tree Search.

\item \textbf{Dealing with the three different sources of uncertainty} present in battles. A well designed AI agent will approach every one differently.

\begin{enumerate}[label=(\roman*)]

\item Players decide simultaneously and often intentionally make unexpected moves. As a result, opponent prediction is critical.

\item There are many kinds of hidden information in the game as many of the opponent Pok\'emon properties are unknown unless revealed or inferred.

\item The damage and the effect of the vast majority of Pok\'emon moves is non deterministic. Given a state, the very same moves can lead to over 1,000 different states, depending on the output of a RNG.

\end{enumerate}

\end{enumerate}

\subsection{Considerations when evaluating states using Machine Learning}

In our opinion, the following are some of the most important considerations one should take into account before applying Machine Learning to this problem.

\begin{enumerate}

\item The game mechanics are very complicated, in contrast to games like GO or Chess. Furthermore, the number of possible battle states in the game is very large, significantly more than the $3^{81}$ possible states in the game of GO. These factors make it hard for Neural Networks to infer patterns in the states that can be exploited. Based on this, directly asking the Neural Networks to evaluate the states is something we do not believe is optimal as it depends, among others, on the Neural Networks learning the very complicated game mechanics indirectly. This approach worked well in games like Chess or GO because their mechanics are very simple.

\item Every Pok\'emon battle can be considered set in an almost unique environment because the teams of the players will almost never be the same, especially in the battle format we focus. As a result, training Neural Networks only using past battles, with simple supervised learning, will likely result in having zero samples of the ongoing battle's environment in the train set. For this reason, we believe it is important for Neural Networks to be trained on the past turns of the ongoing battle too and possibly place more weight on these.

\item The multiple sources of uncertainty in the game make it difficult to determine what role luck played in victory. Given a state, the very same moves can sometimes lead to over 1,000 different states due to the RNG. As a result we believe using a traditional Reinforcement Learning approach where the AI agent is receiving rewards based on the outcome of the battle is not optimal as it is common for players to win even if they play very poorly compared to their opponent and the opposite. In a few cases the team of a player is so much more powerful than the opponent's that the outcome of the battle is determined before it even starts, regardless of the skill of the players.

\end{enumerate}

\subsection{The game trees are pathological}

Over 30 years ago, Dana Nau \cite{b31} \cite{b57} discovered that in some games, looking ahead too many turns leads to worse decision making. This phenomenon was named ``game tree pathology''. Following excessive game analysis, we concluded that the game trees are pathological. To the best of our knowledge, we present the most popular game where this phenomenon is apparent in all game trees. In Pok\'emon, every time we look ahead another turn we have to make assumptions about the decisions of the opponent that turn, guess the hidden information about the properties of the opponent's Pok\'emon and assume the numbers the RNG will generate. The game tree is pathological because the more we look ahead the more we base assumptions on top of assumptions making the AI delusional as it looks into a future far different from what will actually happen. Unfortunately, not looking ahead is not an option in this game. As a result, we have to find the right number of turns to look-ahead in order to see into the future while minimizing the issue we described.

\section{Methods of this paper}

The approaches we describe in this section depend on the conclusions we described above. We first describe our approach to Pok\'emon battling and then we explain in detail how our approach deals with every one of the three different sources of uncertainty present in the game. Finally, we propose an algorithm for team building.

\subsection{A battle algorithm that ensures teamwork}

\subsubsection{The main algorithm}

The algorithm we describe below places a lot of weight into keeping the team of the AI agent balanced while minimizing the issue of game tree pathology, which we discussed in the subsection 3.3. It uses a distributed transposition table to greatly reduce the number of states explored. Further, it relies on a forward pruning algorithm to prune states because alpha beta pruning cannot be used in games like this where players decide simultaneously.

\begin{enumerate}[label=\textbf{Step \arabic*: }]

\item We start from the current battle state and we look-ahead for 2 turns. For both players we consider all possible moves during this process and we end up with several terminal states.

\item For every of the terminal states of step 1 we want to calculate the probability the AI agent will win from that state. This is akin to determining ``how strong is our team compared to the opponent's''. We calculate this in two steps.

\begin{enumerate}[label=\textbf{Step 2.\arabic*: }]

\vspace{1mm}

\item We set each Pok\'emon in our side to battle alone against each Pok\'emon in our opponent's side. In order to determine the outcome of these one-versus-one battles, we rely on 1 or 2 turn look-ahead. The terminal states of this look-ahead are scored with a function. Each Pok\'emon's value at that point is determined based on its remaining life points (HP). It also gets a big bonus if that value is above zero. The scoring function returns the score of our Pok\'emon minus the opponent's Pok\'emon score.

\vspace{2mm}

\item We combine the scores of all one-versus-one battles into a single score by calculating their average. The goal of this is to determine how high chance our team has to win the opponent's. In our current implementation we use a simple average but it would be more appropriate to weight the scores based on the probability the corresponding battle has to occur in the future. We determined that, the more even a battle is, the higher this probability is, since in general players will avoid using Pok\'emon that are weak against the enemy if they have a better choice.

\end{enumerate}

\item Every time the scores of all children of a state are available, in either of the above steps, we construct a payoff matrix. We use this matrix to predict the next move of the opponent at that state, as we describe in the subsection 4.1.2. In that subsection we also describe how the algorithm deals with the other two sources of uncertainty during the look-ahead, namely hidden information and excessive RNG use.

\end{enumerate}

\textbf{A forward pruning approach:} For each player, this algorithm iteratively removes the dominated strategies. To do so it relies on static analysis of the game states. It checks if one of several conditions are met in order to determine if a state is worse than another for the AI agent. For example, if two battle states differ in only one or two parameters, it is not challenging to write pruning conditions for someone familiar with the game. Unlike chess, the unique nature of the game of Pok\'emon makes this pruning approach both very efficient and low risk as it rarely prunes non dominated strategies. To the best of our knowledge, this is the first time such an approach is proposed for this game.

\subsubsection{Dealing with uncertainty}

In this subsection we will describe in detail how the algorithm we presented deals with every one of the three sources of uncertainty we introduced in the section 3.1 (``How battles can be won'').

\begin{enumerate}[label=(\roman*)]

\item \textit{Players decide simultaneously and often intentionally make unexpected moves. As a result, opponent prediction is critical.} 

\vspace{1mm}

At every point in the algorithm we described above we need to predict the move of our opponent we use the following algorithm. Our approach is based on a regret minimization Machine Learning algorithm.

\vspace{2mm}

\begin{enumerate}[label=\textbf{Step \arabic*: }]

\item Our algorithm receives as inputs a payoff matrix as well as the history of opponent moves with the corresponding payoff matrices. Since the game is zero sum, our best move is associated with the largest number in the payoff matrix. The opposite is true for the opponent's best move.

\vspace{2mm}

\item A Machine Learning algorithm takes the inputs we described above and returns the probability of the opponent making each of their available moves.

\vspace{2mm}

\item We multiply the probabilities returned by the Machine Learning algorithm with the payoffs of the possible opponent moves.

\vspace{2mm}

\item From the resulting numbers of the above step, we consider the opponent move associated with the lowest number.

\vspace{2mm}

\item Our algorithm returns as output our best response to the opponent's move we considered in the above step.

\end{enumerate}

\vspace{3mm}

\item \textit{There are many kinds of hidden information in the game as many of the opponent Pok\'emon properties are unknown unless revealed or inferred.}

\vspace{1mm}

We deal with this source of uncertainty with two approaches, depending on whether we know the species property of the opponent's Pok\'emon.

\vspace{2mm}

\begin{enumerate}[label=\textbf{Case \arabic* -}]

\item \textbf{We know the species:} In this case we determine the most probable value for all the unknown properties of the opponent's Pok\'emon in question by relying on detailed usage statistics. During this process we consider the known properties other than species, if any. As the game progresses and more properties are being revealed, our predictions of the unknown properties are getting more and more accurate. In general, the same species do not have extreme variations in the rest of their properties. For this reason this approach is very successful and great human players think this way too.

\vspace{2mm}

\item \textbf{We do not know the species:} In this case all the other properties will be unknown too. We decide with which Pok\'emon we will represent every of the entirely unknown opponent Pok\'emon by solving the following mixed integer linear programming problem using the IBM CPLEX Optimizer \cite{b58}.

\vspace{2mm}

N is the number of all Pok\'emon available in the game, for example, in generation 7 we have N = 802.

M is the number of non fainted Pok\'emon in our team, where 1 $\leq$ M $\leq$ 6.

K is the number of entirely unknown Pok\'emon in the opponent's team, where 0 $\leq$ K $\leq$ 5.

\vspace{2mm}

The objective of the optimization is to select K Pok\'emon such that, the outcome of the hypothetical battle between the M non fainted Pok\'emon in our team and all the N Pok\'emon available in the game is approximated as accurately as possible by the outcome of the hypothetical battle between the M non fainted Pok\'emon in our team and the K selected Pok\'emon.

\begin{align*}
  \min\quad        & \sum_{j=1}^M {| \sum_{i=1}^N x_i   ( s_{ji} - {{1} \over {N}} \sum_{t=1}^N s_{jt} ) |}                                                          \\
  \text{s.t.\quad} & \sum_{i=1}^N x_i = K \\
  & x_i\in\{0,1\}     &    i=1,\dots,N \\
\end{align*}

We define N Boolean decision variables, $x_1, ..., x_N$ where each one corresponds to a specific Pok\'emon from all the available in the game. If  $x_{150} = 1$ this indicates that we select the Pok\'emon with index 150, named Mewtwo, to represent one of the entirely unknown opponent Pok\'emon. Since there are K entirely unknown Pok\'emon in the opponent's team, we must select exactly K Pok\'emon to represent them and thus we set the appropriate constrain.

\vspace{2mm}

The matrix s, given the index of a Pok\'emon in our team and the index of any of the available Pok\'emon in the game, contains a positive or a negative number that represents the approximate outcome of the battle between these two Pok\'emon.  We derive all possible such matrices from a matrix of NxN where the outcome of any Pok\'emon versus any other is stored. We pre compute offline this NxN matrix by having every two Pok\'emon battle each other without teammates for 2 turns.

\end{enumerate}

\vspace{2mm}

\item \textit{The damage and the effect of the vast majority of Pok\'emon moves is non deterministic. Given a state, the very same moves can lead to over 1,000 different states, depending on the output of a RNG.}

\vspace{2mm}

We use the approach we are describing below to deal this source of uncertainty during look-ahead. Our approach achieves much higher accuracy and is significantly more efficient compared to other alternatives. Further, it is deterministic which aids the evaluation of the AI agent. Before we describe our approach we will describe why two obvious alternatives are not a good choice.

\subsubsection*{Determinization}

Starting from a state, considering that the same moves can lead to a very high number of different states, sometimes over 1,000, determinization cannot be considered because it will lead to an extremely high branching factor.

\vspace{1mm}

\subsubsection*{Monte Carlo sampling}

If we have not found a better alternative, Monte Carlo sampling would have been our choice. The primary issue is that it is very inefficient as it requires many samples, and thus a lot of time, to return an accurate approximation of the value of a battle state.

\vspace{1mm}

\subsubsection*{Our approach}

\vspace{1mm}

\begin{enumerate}[label=\textbf{Step \arabic*: }]

\item Each random event in the game requests from the RNG a floating point number in the range [0, 1). We have observed that virtually all events do not nearly use this entire range. For this reason, in our approach we are restricting the RNG to only generate a number among N predefined ones. Following experiments we determined that N should be in range [8, 20]. In the examples below we use N = 8.

\vspace{2mm}

\item The above is not enough on its own because there are usually multiple events that request a random number per turn. Assuming there are five such events and each player has only one available move in a turn, even with our approach, we would need to consider $8^{5}=32,768$ states. We tackle this thanks to two observations. First, the vast majority of random events in the game impact the side of a specific player only. Further, at every turn usually one random event per side has the most impact in the state. The second observation means it will not be a big compromise in terms of accuracy to use the same random number for all events of a side in a turn.

\vspace{2mm}

Following our observations, we are asking the RNG to generate only two numbers per turn. Every time a random event asks for a number, we are proving it with one of the two generated that turn, depending on which side the event will impact. As a result, if both players have only one available move in a turn, we only have to consider $8^{2}=64$ states that turn. Usually players have 4 to 9 moves per turn so we have to consider ${8^{2}}\times{9^{2}}=5,184$ states as they decide simultaneously. In practice this number is significantly lower, around 500, thanks to the use of a transposition table, as we described in our main algorithm.

\vspace{2mm}

\item The approach we described in the above steps has a high branching factor. As a result, we use it in only one turn of look-ahead, the next turn of the battle. In all turns of look-ahead after the next, we force the RNG to always generate 0.5, effectively seeding all these random events with ``average luck''. As the next turn of the battle is usually the most important in look-ahead, little accuracy is lost from this.

\end{enumerate}

\end{enumerate}

\subsection{Team building using a genetic algorithm}

Given a powerful AI agent for battles, a simple and effective team building approach is to rely on a genetic algorithm. In the battles we focus players are provided with random teams, so there was no need for us to implement team building. Regardless, we felt that describing this would make a contribution, as team building was discussed in several papers.

\begin{enumerate}[label=\textbf{Step \arabic* -}]

\item \textbf{Initialization}: In our case, we consider as an individual an entire Pok\'emon team and as a gene a Pok\'emon property. The teams used by human players in the online platform can be very useful if available, as they can be used to seed the initial population of individuals.

\item \textbf{Fitness function}: The fitness function of the genetic algorithm in our case will take as input a Pok\'emon team and evaluate its performance against as many other teams as possible. In order for the fitness function to work well, the teams a candidate team will battle against should represent these used by humans players in the online platform. The evaluation of teams will rely on AI agent self play.

\item \textbf{Selection}: During the selection phase some of the best performing teams are selected.

\item \textbf{Crossover}: During the crossover phase, we generate new teams by taking properties of Pok\'emon from two successful teams.

\item \textbf{Mutation}: During this phase, we sightly modify the properties of Pok\'emon in the newly generated team.

\item \textbf{Termination}: We terminate the algorithm when we find a few teams with a very high fitness score. We test these teams in the online platform, since the fitness function will have a margin of error.

\end{enumerate}

\section{Experiments}

\subsection{Experimental setup}

We connect our AI agent to the online battle platform based on Pok\'emon Showdown using the WebSocket API provided. We play battles against human players in the format ``[Generation 7] Random Battle''. The platform automatically matches the AI agent with opponents based on their ELO rating.

Our AI agent is distributed across 4 Amazon WebServices c5a.24xlarge instances connected to each other using 10 Gbps Ethernet, each featuring a single 48-core AMD EPYC 7R32 processor and 192 GB of main memory. The AI agent must run on high end hardware because the online platform enforces a strict time limit of a few seconds per decision. The most notable distributed component is the transposition table where the servers communicate in an asynchronous and almost entirely peer to peer manner. This approach minimizes delays in decision making due to network delays. Our distributed approach is based on similar approaches developed for Chess \cite{b54} \cite{b55}.

\subsection{Evaluation method}

Evaluating the performance of AI agents in games is critical because it allows researchers to track the progress made, develop AI agents that can learn and improve on their own as well as compare them with each other and human players.

The most popular rating systems by far are the ELO \cite{b32} and Glicko \cite{b33}. These measure the relative skill of a number of players that play against each other. ELO is more popular but Glicko uses ``rating reliability'' that can be used to penalize players that have a long time to play the game. AI agents in Chess and GO are usually compared with the ELO rating but it is incorrect to compare the rating of humans and AI agents directly, unless they have been playing against each other. 

Games that rely on both luck and skill like Pok\'emon present challenges to rating systems. The game of Poker, due to its unique nature, is evaluated with unique approaches. The majority of other popular such games, including video games such as League of Legends, are evaluated with the ELO rating. While there has been research to design rating systems that try to rank players while minimizing the impact of randomness in these rankings \cite{b34}, these systems are not as widely adopted as ELO and Glicko. Game developers, including Nintendo in official games, rely on ELO to rank players. Due to the nature of these games, when evaluating the performance of a player, it is recommended to keep the ELO rating over time, like we do in this work. This would make it possible to determine if luck temporally impacted the rating. If this is the case, there will be big rating spikes or drops.

Pok\'emon Showdown, the platform our AI agent uses to play against humans, ranks players based on the ELO rating \cite{b52}. Glicko and a variation are also displayed. The ladder never resets and the platform makes two modification to the ELO rating system: It adds ladder decay that reduces the ranking of players if they stop playing for some time and it limits the maximum rating gain or loss per battle as the ELO increases. One improvement we suggest to the Pok\'emon Showdown team is to provide, in addition to the existing one, an alternative, seasonal ladder based on the ELO rating without the two modifications we mentioned. That ladder will reset every few weeks or months, exactly like the ladder systems work in many popular games, including the latest generation of official Pok\'emon games. This will make it possible to evaluate AI agents with the pure ELO rating system which is the golden standard used in most game AI agent papers. Furthermore, this will prune from the ladder inactive players, instead of just reducing their ELO rating, which will make calculating the percentiles of players and AI agents more accurate, as we discuss below.

The popularity of different game formats changes over time in the online platform. This is dramatic when a new generation is released, as most players will switch to the newer and the previous will not be as popular. Despite this, a number of very hardcore players will keep playing the previous. As a result, reaching the very top of the ladder does not get noticeably easier but the ELO range becomes much shorter. This in turn means that the best players in the world in the previous generation will have a much lower ELO rating compared to the best players in the new generation, despite having very similar skill level. This observation means that the ELO rating of AI agents can be compared directly if and only if they were evaluated in the exact same battle format and exact same time period. If this is not the case, the only way to compare the AI agents would be via their percentiles. Because it is very unlikely for two AI agents to be evaluated in the exact same time period, we suggest the percentile to be the golden standard in evaluating AI agents in this game. Unfortunately, currently the online platform does not display the rank of players outside the top 500 and thus it is not possible to find the percentile of lower ranked AI agents. We suggest this functionality to be implemented, in addition to the seasonal ladder.

In summary, we believe the percentile is by far the most accurate way to evaluate the performance of Pok\'emon AI agents as long as this information is provided by the platform and not approximated by the researcher. When this is not available, if the AI agent reaches the top 500 players, the world rank can be relied upon. In addition to the world rank, in our evaluation below we provide the ELO rating over time while we clearly specify the battle format and time period the AI agent was evaluated.

\subsection{Results}

\begin{figure}
	\centering
	\subfloat[The world rank of the AI agent after every battle played.\\The platform provides the rank only for the top 500 players.]{{\includegraphics[scale=0.62]{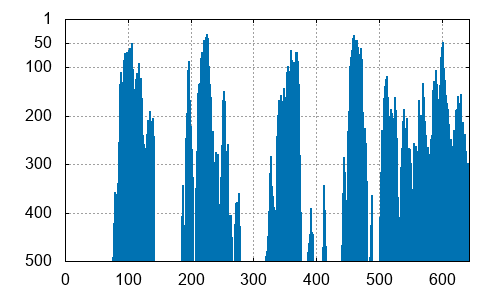}}}
	\vspace{1mm}
	\subfloat[The ELO rating of the AI agent after every battle played.\\Every player starts with ELO 1,000.]{{\includegraphics[scale=0.62]{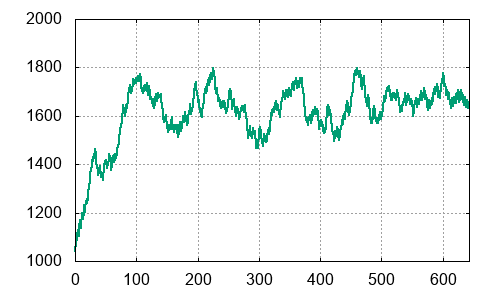}}}
	\caption{The results of our AI agent after 643 battles against human players in the Pok\'emon Showdown platform.}
	\label{fig:pokemon_ai_agent_scores}
\end{figure}

We set our AI agent, named ``Athena2020'', to play 643 battles against human players in the format ``[Generation 7] Random Battle'' in the Pok\'emon Showdown platform. The AI agent was started at approximately 22 Aug 2020 15:03 GMT and was stopped at approximately 25 Aug 2020 17:21 GMT. The online platform automatically matches players to battle against each other based on their ELO rating. Links to the replays of the 643 battles our AI agent played, which are stored at the Pok\'emon Showdown platform, are available in the \hyperref[sec:Appendix]{Appendix}.

In the Figure~\ref{fig:pokemon_ai_agent_scores} (a) the world rank of the AI agent after every battle, is shown. As we discussed, the platform currently provides the rank only for the top 500 players. As it can be seen on the figure, the AI, starting from a new account, first reached the top 500 players after 79 battles. After battle 227 it reached its peak by ranking 33rd in the world. Other notable highs were ranking 52nd after battle 107, 70th after battle 369, 35th after battle 461 and 50th after battle 602. Following a query to the developers of the online platform, we were informed that the total number of players that played at least one battle, in the format the AI was evaluated on, is as of late August 2020 over 15 million. While many of these are inactive and thus we cannot calculate the percentiles over time by diving the ranks over time with this number, we felt it was important to mention it, as it gives an idea of how fierce the competition is on the platform. Since reaching the top 500 players after battle 79, the AI remained ranked among the top 500 players in the world exactly the 2/3rds of the time. We believe this indicates that the performance of the AI agent was not due to mere luck but it truly deserved the rankings it reached. Furthermore, it is obvious that luck does indeed play a role. This supports our claims of high uncertainty in the game.

In the Figure~\ref{fig:pokemon_ai_agent_scores} (b) the ELO rating of the AI agent after every battle played is shown. Every player starts with ELO 1,000. Taking a look at the figure makes apparent that there is a clear correlation between the ELO rating and the rank, as one would expect.  As we discussed above, the ELO rating of AI agents can be compared directly if and only if they were evaluated in the exact same battle format and exact same time period. We provide the ELO rating in order to show the performance of the AI agent when it is not ranked among of the top 500 players.

To the best of our knowledge, this is the only AI agent for Pok\'emon that has managed to reach the top 500 players in the world in any battle format.

This work was ready for publication in late 2020. We planned to spend a few more months to improve the AI's ranking further as there is great potential to do so. However, due to other obligations, we decided to publish this now.

\section{Conclusions}

In this work we explained in depth the mechanics of the Pok\'emon battle system from the perspective of an AI researcher and we performed a game analysis that vastly expanded our understanding of the problem. We continued by proposing unique AI algorithms based on our understanding that the two biggest challenges in the game are keeping a balanced team and dealing with three different sources of uncertainty. Our AI agent performed significantly better than all previous attempts and peaked at the 33rd place in the world, in one of the most popular battle formats, while running on only 4 single socket servers.

\section{Future work}

Our work can be extended to support Double and Triple battles as well as formats other than Random Battles. This last suggestion would require the implementation of the approach we described above titled ``Team building using a genetic algorithm''.  Further, a more sophisticated pruning algorithm can be used. One option is to use the work of Saffidine et al. \cite{b51} where they propose an algorithm inspired by alpha beta pruning for games where players decide simultaneously, since alpha beta pruning itself cannot be used in such games. Another is to develop a new, more suited for our use case, pruning algorithm. Preliminary investigation made us conclude that a pruning algorithm based on the principle of backwards induction could be applicable to the game of Pok\'emon as well as other games and problems.

We also suggest to the Pok\'emon Showdown team to implement the functionality to determine the percentile and rank of a player who is not among the top 500. This would make it possible to compare AI agents of all skill levels even if they were evaluated in different time periods. Furthermore, a seasonal ladder, as we have described above, will aid the evaluation of AI agents.

\section*{Acknowledgments}

The author would like to thank Prof. Ion Androutsopoulos for his suggestions and fruitful discussions about Machine Learning and Game Theory. Furthermore, he is very grateful to Prof. Giannis Marias and Ioannis Sazonof for providing him with access to the computers of the CS Laboratories of Athens University of Economics and Business that helped with the testing of early ideas.

This work would not have been possible without the amazing work of Guangcong Luo, Kirk Scheibelhut and the other developers of the Pok\'emon Showdown online platform and battle simulator.

Finally, special thanks to Nintendo and Game Freak for creating the Pok\'emon franchise.

\bibliographystyle{IEEEtran}
\bibliography{Teamwork_under_extreme_uncertainty_-_AI_for_Pokemon_ranks_33rd_in_the_world}

\begin{thebibliography}{10}
\providecommand{\url}[1]{#1}
\csname url@samestyle\endcsname
\providecommand{\newblock}{\relax}
\providecommand{\bibinfo}[2]{#2}
\providecommand{\BIBentrySTDinterwordspacing}{\spaceskip=0pt\relax}
\providecommand{\BIBentryALTinterwordstretchfactor}{4}
\providecommand{\BIBentryALTinterwordspacing}{\spaceskip=\fontdimen2\font plus
\BIBentryALTinterwordstretchfactor\fontdimen3\font minus
  \fontdimen4\font\relax}
\providecommand{\BIBforeignlanguage}[2]{{%
\expandafter\ifx\csname l@#1\endcsname\relax
\typeout{** WARNING: IEEEtran.bst: No hyphenation pattern has been}%
\typeout{** loaded for the language `#1'. Using the pattern for}%
\typeout{** the default language instead.}%
\else
\language=\csname l@#1\endcsname
\fi
#2}}
\providecommand{\BIBdecl}{\relax}
\BIBdecl

\bibitem{b53}
\BIBentryALTinterwordspacing
Deep blue, IBM's supercomputer, defeats chess champion garry kasparov in 1997.
   \url{https://www.nydailynews.com/news/world/kasparov-deep-blues-losingchess-champ-rooke-article-1.762264}
\BIBentrySTDinterwordspacing

\bibitem{b1}
\BIBentryALTinterwordspacing
D.~Silver, A.~Huang, C.~J. Maddison, A.~Guez, L.~Sifre, G.~van~den Driessche,
  J.~Schrittwieser, I.~Antonoglou, V.~Panneershelvam, M.~Lanctot, S.~Dieleman,
  D.~Grewe, J.~Nham, N.~Kalchbrenner, I.~Sutskever, T.~Lillicrap, M.~Leach,
  K.~Kavukcuoglu, T.~Graepel, and D.~Hassabis, Mastering the game of Go with
  deep neural networks and tree search, \emph{Nature}, vol. 529, no. 7587, pp.
  484--489, Jan 2016.   \url{https://doi.org/10.1038/nature16961}
\BIBentrySTDinterwordspacing

\bibitem{b3}
\BIBentryALTinterwordspacing
A.~Kalose, K.~Kaya, and A.~Kim, Optimal battle strategy in Pok\'{e}mon using
  reinforcement learning.
  \url{https://web.stanford.edu/class/aa228/reports/2018/final151.pdf}
\BIBentrySTDinterwordspacing

\bibitem{b4}
\BIBentryALTinterwordspacing
K.~Chen and E.~Lin, Gotta Train \'Em All: Learning to Play Pok\'{e}mon Showdown
  with Reinforcement Learning.
  \url{http://cs230.stanford.edu/projects_fall_2018/reports/12447633.pdf}
\BIBentrySTDinterwordspacing

\bibitem{b5}
\BIBentryALTinterwordspacing
A Pok\'{e}mon Showdown Battle Bot written in Python.
  \url{https://github.com/pmariglia/showdown}
\BIBentrySTDinterwordspacing

\bibitem{b6}
\BIBentryALTinterwordspacing
AI project for Pok\'{e}mon Showdown.
  \url{https://github.com/taylorhansen/pokemonshowdown-ai}
\BIBentrySTDinterwordspacing

\bibitem{b43}
\BIBentryALTinterwordspacing
Core series Pok\'{e}mon games.
  \url{https://bulbapedia.bulbagarden.net/wiki/Core_series}
\BIBentrySTDinterwordspacing

\bibitem{b44}
\BIBentryALTinterwordspacing
Spin-off Pok\'{e}mon games.
  \url{https://bulbapedia.bulbagarden.net/wiki/Spin-off_Pok%C3\%A9mon_games}
\BIBentrySTDinterwordspacing

\bibitem{b7}
\BIBentryALTinterwordspacing
Bulbapedia - Pok\'{e}mon encyclopedia.
  \url{https://bulbapedia.bulbagarden.net/}
\BIBentrySTDinterwordspacing

\bibitem{b8}
\BIBentryALTinterwordspacing
Smogon - Pok\'{e}mon encyclopedia and battle strategy.
  \url{https://www.smogon.com/}
\BIBentrySTDinterwordspacing

\bibitem{b9}
\BIBentryALTinterwordspacing
Serebii - Pok\'{e}mon encyclopedia and news.   \url{https://serebii.net/}
\BIBentrySTDinterwordspacing

\bibitem{b39}
\BIBentryALTinterwordspacing
S.~Lee and J.~Togelius, Showdown AI Competition, in \emph{2017 IEEE Conference
  on Computational Intelligence and Games (CIG)}.\hskip 1em plus 0.5em minus
  0.4em\relax IEEE Press, 2017, p. 191–198.
  \url{https://doi.org/10.1109/CIG.2017.8080435}
\BIBentrySTDinterwordspacing

\bibitem{b35}
\BIBentryALTinterwordspacing
H.~Ho and V.~Ramesh, Percymon: A Pok\'{e}mon Showdown Artifical Intelligence.
  \url{https://varunramesh.net/content/documents/cs221-final-report.pdf}
\BIBentrySTDinterwordspacing

\bibitem{b40}
\BIBentryALTinterwordspacing
A.~Kalose, K.~Kaya, and A.~Kim, Optimal battle strategy in Pok\'{e}mon using
  reinforcement learning.
  \url{https://web.stanford.edu/class/aa228/reports/2018/final151.pdf}
\BIBentrySTDinterwordspacing

\bibitem{b37}
\BIBentryALTinterwordspacing
M.~L. Sanchez, Learning complex games through self play - Pok\'{e}mon battles.
   \url{https://upcommons.upc.edu/bitstream/handle/2117/121655/134419.pdf}
\BIBentrySTDinterwordspacing

\bibitem{b36}
\BIBentryALTinterwordspacing
L.~Norstr\"{o}m, Comparison of Artificial Intelligence Algorithms for
  Pok\'{e}mon Battles.
  \url{https://odr.chalmers.se/bitstream/20.500.12380/300015/1/Linus\%20Norstr\%C3\%B6m.pdf}
\BIBentrySTDinterwordspacing

\bibitem{b38}
\BIBentryALTinterwordspacing
D.~Huang and S.~Lee, A Self-Play Policy Optimization Approach to Battling
  Pok\'{e}Mon, in \emph{2019 IEEE Conference on Games (CoG)}.\hskip 1em plus
  0.5em minus 0.4em\relax IEEE Press, 2019, p. 1–4.
  \url{https://doi.org/10.1109/CIG.2019.8848014}
\BIBentrySTDinterwordspacing

\bibitem{b48}
\BIBentryALTinterwordspacing
Pok\'emon Bag on Bulbapedia.
  \url{https://bulbapedia.bulbagarden.net/wiki/Bag}
\BIBentrySTDinterwordspacing

\bibitem{b13}
\BIBentryALTinterwordspacing
Pok\'{e}mon species on Bulbapedia.
  \url{https://bulbapedia.bulbagarden.net/wiki/Pok\%C3\%A9mon_(species)}
\BIBentrySTDinterwordspacing

\bibitem{b14}
\BIBentryALTinterwordspacing
List of Pok\'{e}mon species on Bulbapedia.
  \url{https://bulbapedia.bulbagarden.net/wiki/List_of_Pok\%C3\%A9mon_by_National_Pok\%C3\%A9dex_number}
\BIBentrySTDinterwordspacing

\bibitem{b15}
\BIBentryALTinterwordspacing
Pok\'{e}mon level on Bulbapedia.
  \url{https://bulbapedia.bulbagarden.net/wiki/Level\#In_the_core_series}
\BIBentrySTDinterwordspacing

\bibitem{b16}
\BIBentryALTinterwordspacing
Pok\'{e}mon type on Bulbapedia.
  \url{https://bulbapedia.bulbagarden.net/wiki/Type}
\BIBentrySTDinterwordspacing

\bibitem{b17}
\BIBentryALTinterwordspacing
Pok\'{e}mon type chart on Bulbapedia.
  \url{https://bulbapedia.bulbagarden.net/wiki/Type/Type_chart}
\BIBentrySTDinterwordspacing

\bibitem{b18}
\BIBentryALTinterwordspacing
Pok\'{e}mon statistics on Bulbapedia.
  \url{https://bulbapedia.bulbagarden.net/wiki/Statistic}
\BIBentrySTDinterwordspacing

\bibitem{b19}
\BIBentryALTinterwordspacing
Pok\'{e}mon statistics formula on Bulbapedia.
  \url{https://bulbapedia.bulbagarden.net/wiki/Statistic\#Generation_III_onward}
\BIBentrySTDinterwordspacing

\bibitem{b20}
\BIBentryALTinterwordspacing
Pok\'{e}mon Natures on Bulbapedia.
  \url{https://bulbapedia.bulbagarden.net/wiki/Nature}
\BIBentrySTDinterwordspacing

\bibitem{b21}
\BIBentryALTinterwordspacing
Pok\'{e}mon Individual Values (IV) on Bulbapedia.
  \url{https://bulbapedia.bulbagarden.net/wiki/Individual_values}
\BIBentrySTDinterwordspacing

\bibitem{b22}
\BIBentryALTinterwordspacing
Pok\'{e}mon Effort Values (EV) on Bulbapedia.
  \url{https://bulbapedia.bulbagarden.net/wiki/Effort_values}
\BIBentrySTDinterwordspacing

\bibitem{b23}
\BIBentryALTinterwordspacing
Pok\'{e}mon Abilities on Bulbapedia.
  \url{https://bulbapedia.bulbagarden.net/wiki/Ability}
\BIBentrySTDinterwordspacing

\bibitem{b45}
\BIBentryALTinterwordspacing
List of Pok\'emon Abilities on Bulbapedia.
  \url{https://bulbapedia.bulbagarden.net/wiki/Ability\#List_of_Abilities}
\BIBentrySTDinterwordspacing

\bibitem{b24}
\BIBentryALTinterwordspacing
Pok\'{e}mon Items on Bulbapedia.
  \url{https://bulbapedia.bulbagarden.net/wiki/Item}
\BIBentrySTDinterwordspacing

\bibitem{b25}
\BIBentryALTinterwordspacing
List of Pok\'{e}mon Items on Bulbapedia.
  \url{https://bulbapedia.bulbagarden.net/wiki/List_of_items_by_index_number_(Generation_VIII)}
\BIBentrySTDinterwordspacing

\bibitem{b26}
\BIBentryALTinterwordspacing
Pok\'{e}mon Moves on Bulbapedia.
  \url{https://bulbapedia.bulbagarden.net/wiki/Move}
\BIBentrySTDinterwordspacing

\bibitem{b28}
\BIBentryALTinterwordspacing
Pok\'{e}mon damage calculation formula on Bulbapedia.
  \url{https://bulbapedia.bulbagarden.net/wiki/Damage\#Damage_calculation}
\BIBentrySTDinterwordspacing

\bibitem{b27}
\BIBentryALTinterwordspacing
List of Pok\'{e}mon Moves on Bulbapedia.
  \url{https://bulbapedia.bulbagarden.net/wiki/List_of_moves}
\BIBentrySTDinterwordspacing

\bibitem{b29}
\BIBentryALTinterwordspacing
Pok\'{e}mon status conditions on Bulbapedia.
  \url{https://bulbapedia.bulbagarden.net/wiki/Status_condition}
\BIBentrySTDinterwordspacing

\bibitem{b30}
\BIBentryALTinterwordspacing
Pok\'{e}mon Showdown on GitHub.
  \url{https://github.com/smogon/pokemon-showdown}
\BIBentrySTDinterwordspacing

\bibitem{b46}
\BIBentryALTinterwordspacing
Pok\'emon Showdown battle formats.
  \url{https://www.smogon.com/dex/ss/formats/}
\BIBentrySTDinterwordspacing

\bibitem{b47}
\BIBentryALTinterwordspacing
Pok\'emon Showdown OverUsed battle format.
  \url{https://www.smogon.com/dex/ss/formats/ou/}
\BIBentrySTDinterwordspacing

\bibitem{b49}
\BIBentryALTinterwordspacing
Official Pok\'emon GO website.   \url{https://pokemongolive.com/en/}
\BIBentrySTDinterwordspacing

\bibitem{b50}
\BIBentryALTinterwordspacing
Technical Machine Pok\'emon AI.
  \url{https://bitbucket.org/davidstone/technical-machine/src/master/}
\BIBentrySTDinterwordspacing

\bibitem{b59}
\BIBentryALTinterwordspacing
A.~Dotson III, Future Sight AI.
  \url{https://www.pokemonbattlepredictor.com/FSAI}
\BIBentrySTDinterwordspacing

\bibitem{b31}
\BIBentryALTinterwordspacing
D.~S. Nau, M.~Luštrek, A.~Parker, I.~Bratko, and M.~Gams, When is it better
  not to look ahead?, \emph{Artificial Intelligence}, vol. 174, no.~16, pp.
  1323--1338, 2010.
  \url{https://www.sciencedirect.com/science/article/pii/S0004370210001402}
\BIBentrySTDinterwordspacing

\bibitem{b57}
\BIBentryALTinterwordspacing
D.~S. Nau, An investigation of the causes of pathology in games,
  \emph{Artificial Intelligence}, vol.~19, no.~3, pp. 257--278, 1982.
  \url{https://www.sciencedirect.com/science/article/pii/0004370282900029}
\BIBentrySTDinterwordspacing

\bibitem{b58}
\BIBentryALTinterwordspacing
IBM CPLEX Optimizer.   \url{https://www.ibm.com/analytics/cplex-optimizer}
\BIBentrySTDinterwordspacing

\bibitem{b54}
K.~Hasan, A Distributed Chess Playing Software System Model Using Dynamic CPU
  Availability Prediction, 07 2011.

\bibitem{b55}
\BIBentryALTinterwordspacing
C.~Frayn and C.~Justiniano, \emph{The ChessBrain Project --- Massively
  Distributed Chess Tree Search}.\hskip 1em plus 0.5em minus 0.4em\relax
  Berlin, Heidelberg: Springer Berlin Heidelberg, 2007, pp. 91--115.
  \url{https://doi.org/10.1007/978-3-540-72705-7_5}
\BIBentrySTDinterwordspacing

\bibitem{b32}
A.~E. Elo, \emph{The rating of Chessplayers, past and present}.\hskip 1em plus
  0.5em minus 0.4em\relax Ishi Press International, 2008.

\bibitem{b33}
M.~E. Glickman, The glicko system, 1995.

\bibitem{b34}
\BIBentryALTinterwordspacing
P.~Duersch, M.~Lambrecht, and J.~Oechssler, {Measuring skill and chance in
  games}, \emph{European Economic Review}, vol. 127, no.~C, 2020.
  \url{https://ideas.repec.org/a/eee/eecrev/v127y2020ics0014292120301045.html}
\BIBentrySTDinterwordspacing

\bibitem{b52}
\BIBentryALTinterwordspacing
Rating Systems on Pok\'emon Showdown.
  \url{https://www.smogon.com/smog/issue43/elo-hello}
\BIBentrySTDinterwordspacing

\bibitem{b51}
\BIBentryALTinterwordspacing
A.~Saffidine, H.~Finnsson, and M.~Buro, Alpha-Beta Pruning for Games with
  Simultaneous Moves, \emph{Proceedings of the AAAI Conference on Artificial
  Intelligence}, vol.~26, no.~1, pp. 556--562, Sep. 2021.
  \url{https://ojs.aaai.org/index.php/AAAI/article/view/8148}
\BIBentrySTDinterwordspacing

\end{thebibliography}

\section*{Appendix}
\label{sec:Appendix}

\subsection*{A. Battle Replays}

\url{http://sarantinos.me/pokemon_ai_battle_replays}

\end{document}